\pdfoutput=1

\documentclass[11pt]{article}

\usepackage{EMNLP2023}

\usepackage{times}
\usepackage{latexsym}

\usepackage[T1]{fontenc}

\usepackage[utf8]{inputenc}
\usepackage{CJKutf8}
\usepackage{microtype}

\usepackage{inconsolata}
\usepackage{graphicx}
\usepackage{subfigure}
\usepackage{multirow}
\usepackage{makecell}

\title{Generative Annotation for ASR Named Entity Correction}

\author{Yuanchang Luo\textsuperscript{\rm}\footnotemark[1],
    Daimeng Wei\textsuperscript{\rm}\footnotemark[1],
    Shaojun Li\textsuperscript{\rm}\footnotemark[1],
    Hengchao Shang\textsuperscript{\rm},
  Jiaxin Guo\textsuperscript{\rm},\\
  \bf{Zongyao Li\textsuperscript{\rm},}
  \bf{Zhanglin Wu\textsuperscript{\rm},}
  \bf{Xiaoyu Chen\textsuperscript{\rm},}
  \bf{Zhiqiang Rao\textsuperscript{\rm},}
  \bf{Jinlong Yang\textsuperscript{\rm},}
  \bf{Hao Yang\textsuperscript{\rm}}\\
  \textsuperscript{\rm}Huawei Translation Service Center, Beijing, China\\
  \tt \{luoyuanchang1,weidaimeng,lishaojun18,shanghengchao,guojiaxin1,lizongyao, \\
  \tt wuzhanglin2,chenxiaoyu35,raozhiqiang,yangjinlong7,yanghao30\}@huawei.com \\
  }

\begin{document}
\maketitle
\renewcommand{\thefootnote}{\fnsymbol{footnote}}
\footnotetext[1]{These authors contributed equally to this work.}
\renewcommand{\thefootnote}{\arabic{footnote}}
\begin{abstract}
End-to-end automatic speech recognition systems often fail to transcribe domain-specific named entities, causing catastrophic failures in downstream tasks. Numerous fast and lightweight named entity correction (NEC) models have been proposed in recent years. These models, mainly leveraging phonetic-level edit distance algorithms, have shown impressive performances. However, when the forms of the wrongly-transcribed words(s) and the ground-truth entity are significantly different, these methods often fail to locate the wrongly transcribed words in hypothesis, thus limiting their usage. We propose a novel NEC method that utilizes speech sound features to retrieve candidate entities. With speech sound features and candidate entities, we inovatively design a generative method to annotate entity errors in ASR transcripts and replace the text with correct entities. This method is effective in scenarios of word form difference. We test our method using open-source and self-constructed test sets. The results demonstrate that our NEC method can bring significant improvement to entity accuracy. The self-constructed training data and test set is publicly available at github.com/L6-NLP/Generative-Annotation-NEC.
\end{abstract}

\section{Introduction}

End-to-end automatic speech recognition (ASR) systems \cite{Graves_Jaitly_2014,6af3452a28a04980b2b8f5eb48730d36,Graves_2012} achieve significant improvements in recent years and the wide usage of weak supervised \cite{radford2022robust} and unsupervised \cite{zhang2023google} data further improves ASR performance. SOTA ASR models achieve considerably low word error rate (WER) on open-source ASR test sets, such as GigaSpeech \cite{chen2021gigaspeech} or LibriSpeech \cite{7178964}.
However, they often mistranscribe domain-specific words, such as person names, locations or organizations, into common words, causing severe misunderstanding.
\begin{figure}
\centering
\includegraphics[height=3.7cm,width=8.0cm]{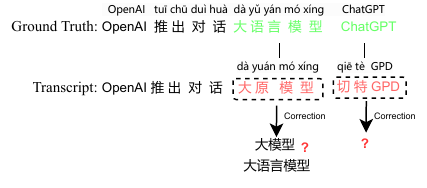}

\caption{The drawback of NEC methods based on phonetic-level similarity algorithms in scenarios when the word form of the ground-truth entity is greatly different from that of the to-be-corrected text.}

\label{figure:vanER}
\end{figure}

In recent years, numerous works \cite{pundak2018deep,jain2020contextual, Wang2020ASREC,le2021contextualized} propose NEC methods to correct named entity errors in ASR transcripts. We divide these methods into two categories: (1) correct errors along with transcript generation; and (2) correct errors after transcript generation, namely, post-editing errors. In category (1), a number of methods \cite{8682441,Garg2020,huber2021instant,wang2023improving} train additional modules to equip ASR models with the capability of contextual bias. Other methods \cite{guo2019spelling,zhang-etal-2020-spelling,zhang2019automatic,Ma2023NbestTR} directly use pre-trained models \cite{devlin2019bert,liu-etal-2020-multilingual-denoising} of text to correct errors in transcripts. Methods in category (1) require modifications to ASR systems in order to equip ASR systems the capability of error correction, so these methods can hardly be applied to third-party ASR systems. 



In contrast, methods in category (2) require no modification to ASR systems, so post-editing NEC methods are more applicable, especially when using ASR systems that are running in the cloud. Recent works under this category focus on solving issues like slow inference speed and lack of phonetic constraints due to the use of non-autoregressive models \cite{leng2022fastcorrect,leng2022fastcorrect2,leng2023softcorrect}. 

Among those, fast and lightweight methods based on text and phonetic-level similarity computed by edit distance algorithm have shown significant performance \cite{raghuvanshi-etal-2019-entity,Garg2020} 
(we refer this method as PED-NEC hereinafter). 

However, although this method is simple and effective, its performance deteriorates greatly in scenarios when there is a great difference between the word forms of the ground-truth entity and the to-be-corrected text. When the forms of entity and related incorrect text in ASR transcripts are similar, we can easily locate mistakes by traversing entity datastore. However, when the forms are different, it is hard to locate the to-be-corrected words by simply traversing the ground-truth entity datastore. 


As shown in Figure \ref{figure:vanER}, the Chinese ASR model mistakenly transcribes "\begin{CJK}{UTF8}{gbsn}{大语言模型}\end{CJK}" (large language model) as "\begin{CJK}{UTF8}{gbsn}{大原模型}\end{CJK}" (large original model). Methods based on text and phonetic-level edit distance have difficulties to determine whether the correct entity is "\begin{CJK}{UTF8}{gbsn}{大模型}\end{CJK}" (large model) or "\begin{CJK}{UTF8}{gbsn}{大语言模型}\end{CJK}" (large language model), because the word form of the incorrect content is different from the correct entity. This issue is especially common for loanwords and entities that contain digits. For example, a Chinese ASR system transcribes "ChatGPT" as "\begin{CJK}{UTF8}{gbsn}{切特GPD}\end{CJK}", making it particularly challenging for NEC methods that are based on phonetic similarity search.

To address the issue above mentioned, we innovatively propose an NEC method using a generative approach to annotate to-be-corrected text in transcript. To be more specific, we utilize speech sound feature, candidate named entity, and ASR transcript to generate (label) to-be-corrected words in the transcript, and perform correction accordingly. This NEC method, which is based on error annotation, achieves end-to-end text correction after identifying the to-be-corrected text, without the need to consider word form changes, so it is superior to previous rule-based replacement approaches. 

We validate the effectiveness of our method on both open-source Aishell \cite{8384449} test sets and self-constructed BuzzWord set, and results show that our method outperforms PED-NEC. Particularly, our method significantly outperforms the PED-NEC method when the word forms of the to-be-corrected text and correct entity are different, as well as on our challenging BuzzWord test set.

\section{Method}
The rationale of PED-NEC is that ASR systems often mistranscribe entities to phonetically similar common words. PED-NEC is a two-step approach: (1) entity retrieval based on speech sound similarity and (2) text correction. Compared to PED-NEC, our method replaces step (1) with direct use of audio for retrieval, which we believe helps solve NEC errors such as "\begin{CJK}{UTF8}{gbsn}{切特GPD}\end{CJK}". Then we employ a generative approach for text correction.

\begin{figure*}
\centering
\includegraphics[height=5cm,width=16.1cm]{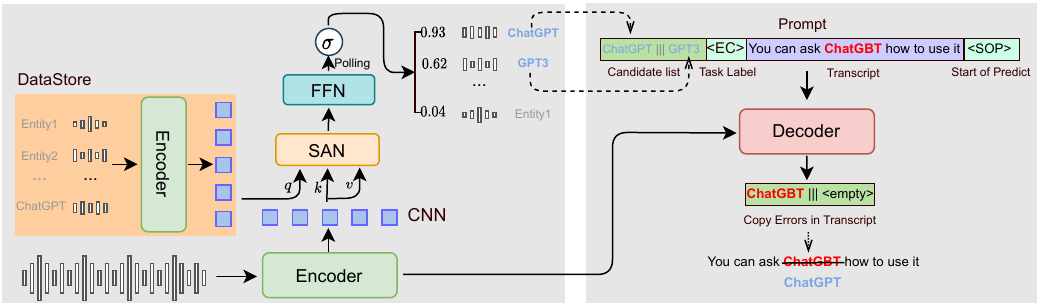}

\caption{Our method consists of two steps: The left part (SS) denotes datastore construction and candidate entity retrieval. The right part (GL) denotes concatenating candidate entities and ASR transcript as a prompt to guide model generate errors in the transcript. Finally, error correction is done by text replacement.}
\label{figure:ECmain}
\end{figure*}

Our method is based on a pre-trained Attention-based Encoder-Decoder (AED) ASR system. The correction process is shown in Figure \ref{figure:ECmain}. A datastore is constructed in advance to store audio-text pairs of entities. After the speech segment and ASR transcript are obtained, speech retrieval is performed to determine whether a part of the speech segment shares similar speech sound features with any candidate entity in the datastore. If yes, we then concatenate the candidate entity and the ASR transcript as a prompt to guide the correction model to generate the possible wrong word(s) in the ASR transcript corresponding to the correct entity. Finally, we replace the wrong text with the correct entity in the datastore. We detail the process of each step in the following part of this section.

\subsection{Datastore Creation}
For the list of entities $X=\{x_1,x_2,...x_n\}$ we collected, we can obtain their speech sounds \begin{equation}
Speech_{x_i}=TTS(x_i)
\end{equation}
via text-to-speech (TTS) engine. Then we input the TTS-generated audios to encoder, and use the output of the last layer of the encoder as the phonetic representation of the entity $x_i$. To improve retrieval accuracy and reduce memory usage, we add a Convolutional Neural Network (CNN) layer to the end of the encoder. So the audio representation of entity $x_i$ is denoted as:
\begin{equation}
   x_i'=CNN(Encoder(Speech_{x_i})) 
\end{equation}

As a result, the datastore stores key-value (representation-entity) pairs 
\begin{equation}
\{(x'_1,x_1),(x'_2,x_2),...(x'_i,x_i)...\}
\end{equation}

\subsection{Entity Retrieval}
We then input the speech segment $s$ to the encoder and get its representation $s’$ from the output of the last layer of the encoder: 
\begin{equation}s'=CNN(Encoder(s))\end{equation}

We introduce self-attention network (SAN) and feed-forward network(FFN) to calculate the probability $p_i$ that $s$ contains a candidate entity $x'_i$ in the datastore. The probability is denoted as: 
\begin{equation}p_i=Sigmoid(FFN(SAN(q=x'_i;k,v=s')))\end{equation}

It should be noted that the input SAN and q are representations of $x'_i$. $k$ and $v$ are key and value of candidate entity $s'$. In addition, we apply average polling after FFN for final classification.

Finally, we obtain the probabilities 
\begin{equation}
    \{p_1,p_2,...p_i...\} 
\end{equation} of whether any entity in the datastore is in the speech segment. We select top K candidate entities for further correction if the probability $p_i$ is higher than the threshold we set. 

\subsection{Error Correction}
We obtain several candidate entities through entity retrieval as described above. As shown in Figure \ref{figure:ECmain}, we concatenate entities with symbol "|||" and then concatenate the entity string with ASR transcript using "<EC>". The entity+transcript string is used as a prompt to guide the correction model generate wrong entities in the transcript that share similar sound features as the candidiate entity. The process is actually a generative annotation method as the correction model outputs one or several words in the original ASR transcript. 

Our generative method is insensitive to word form difference between the to-be-corrected text and candidate entity, thereby solving the issue described in Figure \ref{figure:vanER}.


\begin{table}
\centering
\begin{tabular}{l|c}
\hline
 Type & Predict Errors\\
\hline
1  & <empty> ||| <empty> ||| Error3 \\ 
2  & Error1 ||| Error2 ||| Error3 \\ 
3  & Error1-1,Error1-2 ||| <empty> ||| Error3 \\
\hline
\end{tabular}

\caption{Several possible forms of prediction errors when there are three candidate entities.}
\label{tab:ECform}
\end{table}

In addition, our method also possesses the capability of Entity Rejection. If the model cannot match a candidate entity with a possible wrong entity in the transcript, it will generate symbol "<empty>" to indicate no result is returned. 

We believe this method can easily identify the to-be-corrected text, as it combines the original audio, the candidate entity, and the incorrect transcript. The model aims to find the to-be-corrected text that shares similar speech sounds and aligns with language model. The final step is to replace wrong text with the ground-truth entity in the datastore. 


Using a generative approach to predict incorrect text, we can easily handle various error correction scenarios. As shown in Table \ref{tab:ECform},  where three candidate entities are retrieved, the returned result from the correction model may have different formats. If a candidate entity does not match any piece of text in transcript, an "<empty>" symbol is returned to skip correction. In addition, when a candidate entity matches more than one mistake (type 3 in Table \ref{tab:ECform}), our method can correct all of them.

\section{Experimental Setup}
\label{sec:ExperimentalSetup}

\begin{figure}
\centering
\includegraphics[height=5cm,width=8.0cm]{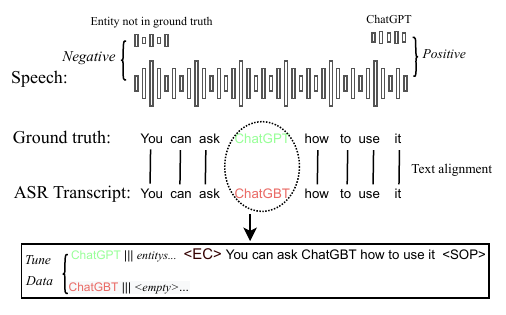}

\caption{Constructing generative labeling training data using speech with ground-truth transcript.}
\label{figure:DataCreation}
\end{figure}

\subsection{Training Data}
To train the correction model, labeled entities in the ground-truth transcripts are required. Thanks to \citet{chen2022aishell} and \citet{yadav2020end}, we obtained 54,129 Chinese entities in Aishell dataset. We refer to their labeling framework to construct our training data. Audio-text pairs that contain labeled entities are used as positive samples while pairs with no entity are treated as negative samples (ten times the number of positive samples). Speech sounds for entities are generated via TTS\footnote{https://github.com/espnet/espnet}.

As shown in Figure \ref{figure:DataCreation}, to equip the pre-trained model with error correction capability, the pre-labeled entity data mentioned above is used to construct fine-tuning data. We first use the Whisper-base model to generate ASR transcripts that may contain incorrect entities, and align them with correct ones using edit distance. The amount of fine-tuning data is less than the data used for training the classification model.

We only use 10k training data. To enable the model to generate "<empty>" when no correction is needed, 20\% prompts contain entities that are not in the transcript, or only partly correct (for example, if the entity that needs to be corrected is "\begin{CJK}{UTF8}{gbsn}{文心一言}\end{CJK}", the entity in our prompt might be "\begin{CJK}{UTF8}{gbsn}{文心言}\end{CJK}", thus the expected result is "<empty>").


It should be noted that all of our training data can be automatically constructed based on the current open-source data, making it easy for other researchers to reproduce our experiments.

\begin{table*}
\centering
\setlength{\tabcolsep}{1.2mm}{
\begin{tabular}{l|cccc|cccc}
\hline
 & \multicolumn{4}{c}{\textbf{AISHELL Test Set (\%)}} & \multicolumn{4}{c}{\textbf{Word Form Variation Set (\%)}}\\
\cline{2-9}
 \textbf{Model} & CER$\downarrow$ & NNE CER$\downarrow$ & NE CER$\downarrow$ & NE Recall$\uparrow$ & CER$\downarrow$ & NNE CER$\downarrow$ & NE CER$\downarrow$ & NE Recall$\uparrow$ \\

\hline
Whisper & 10.47 & \textbf{10.00} & 15.41 & 70.85 & 18.99 & \textbf{18.10} & 25.34 & 25.4  \\
\hline
PED-NEC & 10.40	& 10.42	& 10.85	& 83.34 & 17.60	& 18.32	& 13.58	& 50.79   \\
\hline
PED+GL &10.00	&10.00	&10.03	&84.34	&16.76	&18.12	&12.55	&50.85  \\
SS+NEC &10.26	&10.41	&8.34	&86.20	&17.01	&18.19	&13.05	&50.82  \\
SS+GL &\textbf{9.85}	&10.01	&\textbf{7.41}	&\textbf{87.31}	&\textbf{11.45}	&18.10	&\textbf{7.53}	&\textbf{86.51}  \\
\hline
\end{tabular}}

\caption{Our error correction results on the Aishell test set and the Word Form Variation Set we constructed.}
\label{table:mainres}
\end{table*}

\begin{table}
\centering
\begin{tabular}{l|cccc}
\hline
 & \multicolumn{4}{c}{\textbf{BuzzWord  Test Set (\%)}}\\
\cline{2-5}
 \textbf{Model} &  & NNE & NE  & NE\\
 & CER$\downarrow$ & CER$\downarrow$ &CER$\downarrow$ &  Recall$\uparrow$\\
\hline
Whisper & 16.23 & \textbf{15.29} & 46.49 & 12.22   \\
\hline
PED-NEC & 10.67	& 15.49	& 23.62	& 61.82  \\
\hline
PED+GL &15.00	&15.29	&12.9	&79.96	\\
SS+NEC &16.01	&15.47	&17.40	&70.03	\\
SS+GL &\textbf{14.77}	&15.29	&\textbf{7.26}	&\textbf{87.47}	 \\
\hline
\end{tabular}

\caption{The experiment results of our error correction method on the BuzzWord test set.}
\label{table:buzzword}
\end{table}

\subsection{Test Set}

We use two test sets to verify the effectiveness of our NEC method. One is the Aishell test set, and the other is the BuzzWord test set that we constructed.

We merge all the deduplicated NEs (a total of 3,101) from both the dev and test sets of Aishell to serve as the NE database for the Aishell test set.
To better demonstrate the effectiveness of our method in challenging scenarios, we construct a BuzzWord test set.

Some of these buzzwords are long entities, loanwords, or entities consisting of digits, which are really challenging to ASR systems. The word forms of these words transcribed by ASR systems often vary greatly to that of the ground-truth buzzwords.


The BuzzWord test set contains 1500 short speech segments and corresponding ground-truth transcripts from January 2023 to January 2024. In the test set, we construct 500 positive test cases that contains buzzwords and 1000 negative test cases without buzzwords. 

To make our test set more close to real error correction scenarios, we take speech diversity into consideration. For each buzzword, we collect 10 positive test cases from at least 5 speakers, and we carefully balance female and male voices. Negative samples are also from those speakers. These buzzwords appear at the beginning, in the middle, or at the end of the speech segment, and a buzzword may appear more than once in one speech segment. For details about the buzzwords test set, see Appendix Table \ref{tab:testset}.


Although we only have 50 buzzwords, our experiment shows that this test set poses a great challenge to existing ASR systems.



\subsection{Evaluation Metrics}
Followed by \citet{wang2024dancer}'s work, we assess the performance of various NEC methods using four key metrics:
\begin{itemize}
\item \textbf{CER}: measures the total character error rate of the entire test set.
\item \textbf{NNE-CER}: evaluates the character error rate for characters within the utterance that do not form part of an entity.
\item \textbf{NE-CER}: determines the character error rate for characters that constitute entities within the utterance.
\item \textbf{NE-Recall}: gauges the recall rate of entities within the utterance that are accurately recognized.
\end{itemize}

\subsection{Parameters}
The ASR AED pre-trained model we used is Whisper-base\footnote{https://github.com/openai/whisper}. In speech classification, we use a one-dimensional CNN with a window size of 3 and a stride of 2. The dimension of the SAN is 512, and the hidden layer dimension of FFN is 2048. During training, we use one GPU, with a batch size of 512 and a learning rate of 5e-5. We use a constructed dev set to determine the convergence of the model. The encoder parameters of the pre-trained model are frozen during training and fine-tuning. During fine-tuning, the batch size is set to 64 and the learning rate to 1e-4. During entity retrieval, we select a candidate entity as prompt if the probability is greater than 0.3, with a maximum of 5 candidate entities in one speech segment.


\subsection{Baseline System}

The ASR results for all test sets are generated by Whisper, which is trained on a large amount of weakly supervised data. We used Whisper-large v2\footnote{https://github.com/openai/whisper} in our experiment. For system comparison, we focus on the method based on Phonetic-level Edit Distance \cite{raghuvanshi-etal-2019-entity}, namely the previously mentioned PED-NEC, as a strong baseline. Our method use the same implementation method as \citet{wang2024dancer}\footnote{https://github.com/Amiannn/Dancer}, which additionally includes a preliminary Corrupted Entity Detection (CED) module. The implementation details of the baseline are described in Appendix \ref{expdetails}. We also test our method on commercial ASR systems like iFlytek\footnote{https://www.xfyun.cn/services/lfasr} and Amazon\footnote{https://aws.amazon.com/transcribe} on the BuzzWord test set.
\vspace{10pt}
\section{Result}

In addition to comparing with \textbf{PED-NEC}, our method has three different variants. \textbf{PED+GL} is to find candidate results using PED, and then correct them using our generative annotation method (GL). \textbf{SS+NEC} determines whether a speech segment contains a certain entity based on the entity speech sound and the input speech segment (SS), and then applies the PED-NEC approach for correction. \textbf{SS+GL} is shown in Figure \ref{figure:ECmain}. Which is to find candidate results using SS and then correct them using GL.

We verify the effectiveness of our method on the Aishell and self-constructed BuzzWord test sets. On the Aishell test set, we specifically compare performances of different NEC methods in scenario when the word form of the to-be-correct text is different from the word form of the candidate entity. In addition, we also test our method upon commercial ASR systems to demonstrate generalizability of our method (Appendix Table \ref{table:buzzwordcommercial})


\begin{table*}
\centering
\begin{tabular}{c|c}
\hline
No. & Result \\
\hline 
1 & \makecell*[l]{\begin{CJK}{UTF8}{gbsn}{Ref: 到上世纪50年代后\textcolor{green}{长江白鲟}(cháng jiāng bái xún)就只分布于长江及出海口}\end{CJK}\\ 
\begin{CJK}{UTF8}{gbsn}{ASR: 到上世纪50年代后\textcolor{red}{长江白旭云}(cháng jiāng bái xù yún)就只分布于长江及出海口}\end{CJK} \\ 
\begin{CJK}{UTF8}{gbsn}{PED-NEC:到上世纪50年代后\textcolor{red}{蓝箭白旭云}(lán jiàn  bái xù yún)就只分布于长江及出海口}\end{CJK} \\
\begin{CJK}{UTF8}{gbsn}{Ours: 到上世纪50年代后\textcolor{green}{长江白鲟}(cháng jiāng bái xún)就只分布于长江及出海口}\end{CJK} \\
\begin{CJK}{UTF8}{gbsn}{Explanation: The ASR system wrongly treats the word "鲟 (xún)" as a linking pronunciation}\end{CJK} \\
\begin{CJK}{UTF8}{gbsn}{of two words "旭云 (xù yún)", and thus mistranscribes the word.}\end{CJK} \\}\\
\hline
2& \makecell*[l]{\begin{CJK}{UTF8}{gbsn}{Ref: \textcolor{green}{华硕灵耀}(huá shuò líng yào)X双屏Pro在外观设计还是性能上都有着非常高的水准}\end{CJK}\\ 
\begin{CJK}{UTF8}{gbsn}{ASR: \textcolor{red}{华硕01}(huá shuò líng yāo)X双屏Pro在外观设计还是性能上都有着非常高的水准}\end{CJK} \\ 
\begin{CJK}{UTF8}{gbsn}{PED-NEC: \textcolor{green}{华硕灵耀}\textcolor{red}{01}X双屏Pro在外观设计还是性能上都有着非常高的水准}\end{CJK} \\
\begin{CJK}{UTF8}{gbsn}{Ours: \textcolor{green}{华硕灵耀}X双屏Pro在外观设计还是性能上都有着非常高的水准}\end{CJK}\\
\begin{CJK}{UTF8}{gbsn}{Explanation: A mistranscription of Chinese words "灵耀 (líng yào)" to numbers "01 (líng yāo)"}\end{CJK} \\}\\
\hline
3 & \makecell*[l]{\begin{CJK}{UTF8}{gbsn}{Ref: \textcolor{green}{Midjourney}真的是一个非常非常棒的这个绘图软件}\end{CJK}\\ 
\begin{CJK}{UTF8}{gbsn}{ASR: \textcolor{red}{米德仲尼}(mǐ dé zhòng ní)真的是一个非常非常棒的这个绘图软件}\end{CJK} \\ 
\begin{CJK}{UTF8}{gbsn}{PED-NEC: \textcolor{red}{米德仲尼}(mǐ dé zhòng ní)真的是一个非常非常棒的这个绘图软件}\end{CJK} \\
\begin{CJK}{UTF8}{gbsn}{Ours: \textcolor{green}{Midjourney}真的是一个非常非常棒的这个绘图软件}\end{CJK}  \\
\begin{CJK}{UTF8}{gbsn}{Explanation: A mistranscription of English word "Midjourney" to Chinese words "米德仲尼}\end{CJK} \\
\begin{CJK}{UTF8}{gbsn}{(mǐ dé zhòng ní)"}\end{CJK}\\}\\

\hline
\end{tabular}
\caption{Examples of comparing PED-NEC and our method when the word form of transcribed entity results and the word form of the entity are different.}
\vspace{-10pt}
\label{table:casestudy}
\end{table*}

\subsection{Aishell Result}

Experiment results are shown in Table \ref{table:mainres}. On the AISHELL test set, Whisper already achieves a relatively high accuracy in terms of NE transcription, with a Recall of 70.85\%. Our baseline error correction method, PED-NEC, further increases the Recall to 83.34\% upon Whisper. The improvement is significant, demonstrating PED-NEC is an effective method. However, it should be noted the PED-NEC slightly increase NNE-CER, indicating that this method has a tendency of over-correction.


When we use PED for entity retrieval and GL for correction (PED+GL), we observe improvements on all four metrics, with an increase of more than one point in terms of entity recall. However, NNE-CER achieves similar performance as the baseline, indicating that over-correction is rare when GL is used. When the entity retrieval employs the SS method, but error correction still uses the PED-NEC approach (SS+NEC), we observe varying degrees of improvement in both NE-CER and NE-Recall. However, both CER and NNE-CER decrease, indicating that the SS-based entity retrieval method performs better but still fails to address cases of over-correction. Our proposed SS+GL method gets the lowest CER (9.85) and highest NE recall (87.31\%). And NNE-CER is about 10, which is very close to the best result.


We construct a Word Form Variation set by manually selecting 50 NEs from the AiShell test set of which the word forms of the incorrect text and the ground-truth entity are different (some word form changes are due to the addition of punctuation marks). On this test set, we find that our method significantly outperforms PED-NEC.
\subsection{BuzzWord Result}

A majority of the entities in our BuzzWord test set are newly-created words from January 2023 to January 2024, so most of them are OOVs to ASR systems. In addition, many of the entities are combinations of Chinese characters, English letters, and digits. As the word form of the incorrect text generated by ASR system often differs from that of the ground-truth entity, this test set is challenging for entity retrieval and correction. 


As shown in Table \ref{table:buzzword}, the NE-Recall of Whisper is only 12.22\%, indicating correction of these buzzwords is urgently required. Although PEC-NEC remains effective, its NE-Recall is only 61.82\%. However, when PEC-NEC is used along with our proposed GL, the best NE-Recall can reach 79.96\% while we observe different levels of improvements regarding other metrics. The reason is that our proposed GL is capable of deciding when no correction is required. Moreover, the PED+GL significantly outperforms the SS+NEC, which also demonstrates the superiority of the GL. Our method is much more noise tolerant and the correction performance is not compromised. We discuss this capability in detail in section \ref{entityreject}. Our SS+GL method achieves the highest NE-Recall (87.47\%), a 26\% improvement over PED-NEC, and the lowest CER (14.77), demonstrating its effectiveness.


\begin{figure*}
\centering
\includegraphics[height=7.5cm,width=16cm]{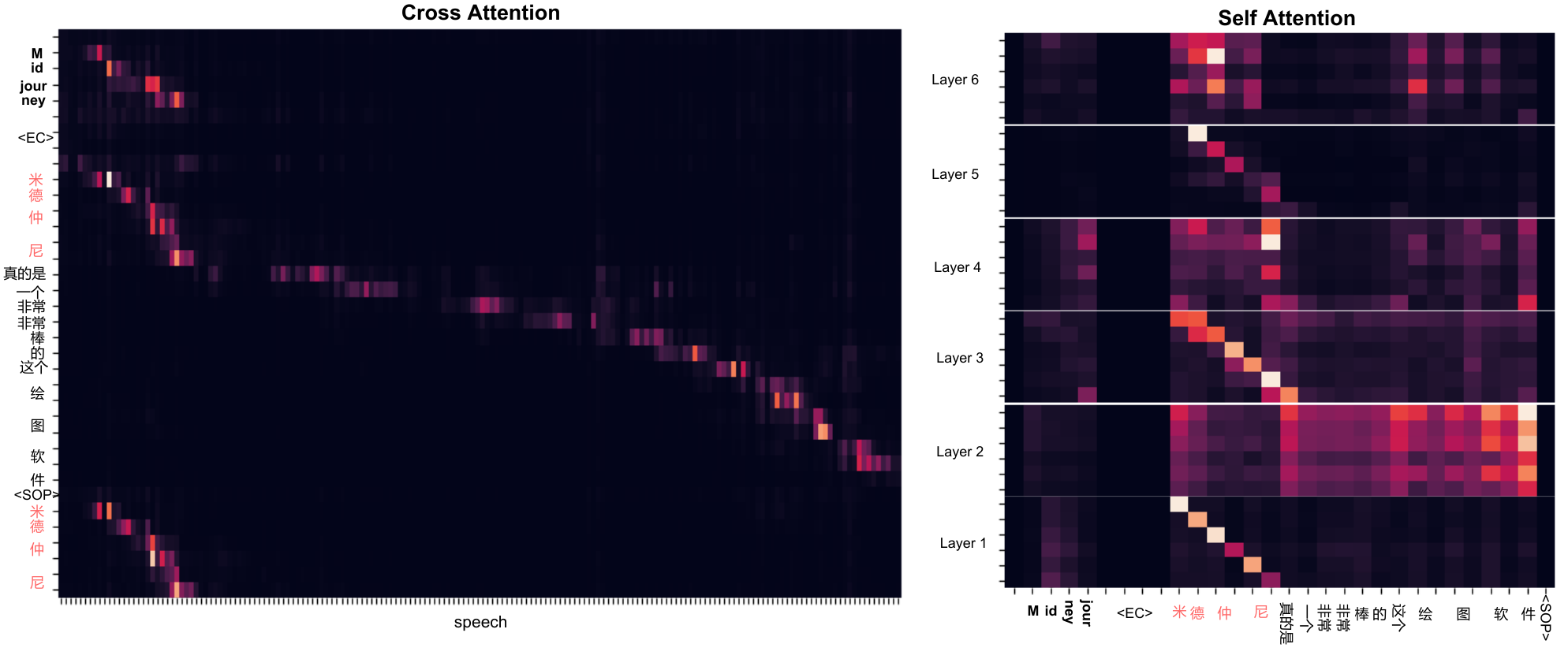}

\caption{Heatmaps of Cross Attention in the last layer and Self Attention in each layer of our generative annotation model. Regarding Self Attention, we analyze the relationship between the output result "\begin{CJK}{UTF8}{gbsn}{米德仲尼}\end{CJK}" and the prompt. The candidate entity is "Midjourney," the incorrectly transcribed text is "\begin{CJK}{UTF8}{gbsn}{米德仲尼}\end{CJK}", and the annotation result is "\begin{CJK}{UTF8}{gbsn}{米德仲尼}\end{CJK}".}
\vspace{-10pt}
\label{figure:heatmap}
\end{figure*}
%
\subsection{Case Study}

As shown in Table \ref{table:casestudy}, we list some cases when PED-NEC fails to correct entities due to word form difference between the to-be-corrected text and ground-truth entity, while our method succeeds.


Regarding case No.1, ASR system transcribes "\begin{CJK}{UTF8}{gbsn}{长江白鲟}\end{CJK}" (Yangtze River Chinese Sturgeon) as "\begin{CJK}{UTF8}{gbsn}{长江白旭云}\end{CJK}", where the to-be-corrected text is longer. PED-NEC mis-corrects part of the entity "\begin{CJK}{UTF8}{gbsn}{长江}\end{CJK}" to a totally wrong entity "\begin{CJK}{UTF8}{gbsn}{蓝箭}\end{CJK}". Our method, however, precisely annotates the to-be-corrected text and replaces it with the ground-truth entity. Regarding case No.2, ASR system transcribes "\begin{CJK}{UTF8}{gbsn}{华硕灵耀}\end{CJK}" (Asus Lingyao) as "\begin{CJK}{UTF8}{gbsn}{华硕01}\end{CJK}", turning part of the Chinese characters into numbers, which is a very tricky case for correction. PED-NEC fails to identify the entity boundary and leaves the digits uncorrected, but our method makes a correct replacement. Regarding case No.3, ASR system transcribes the English entity "Midjourney" as Chinese characters "\begin{CJK}{UTF8}{gbsn}{米德仲尼}\end{CJK}". PED-NEC fails to make a replacement but our method again performs well. 


\section{Analysis}
\subsection{Joint Annotation}

To better analyze the roles of speech segment, candidate entity and ASR transcript in error annotation, we check the cross attention of ASR transcript and speech segment, as well as the self-attention of prompt. As shown in Figure \ref{figure:heatmap}, to analyze the cross attention, we trim speech audios to segments that align with the transcripts. We use the average value of each audio frame and text to denote the cross attention. 
 

As expected, the text (\begin{CJK}{UTF8}{gbsn}{"米德仲尼"}\end{CJK}) generated by the annotation model, the candidate entities ("Midjourney"), and the to-be-corrected text (\begin{CJK}{UTF8}{gbsn}{"米德仲尼"}\end{CJK}) in the transcript all have high attention values with the same segment of the speech signal. Similarly, we analyze the relationship between the annotation result and the prompt. We find that the annotation result pays a lot of attention to the candidate entity and the corresponding to-be-corrected text (although performances vary at each layer). The cross-attention and self-attention heatmap again corroborate our previous hypothesis. We believe this approach is able to accurately annotate the to-be-corrected text that shares similar speech sounds to the candidate entity. This approach remains effective when the word form of the to-be-corrected words is different from that of the candidate entity.  


\begin{figure}
\centering
\setlength{\belowcaptionskip}{-0.4cm}
\includegraphics[height=5.6cm,width=7.8cm]{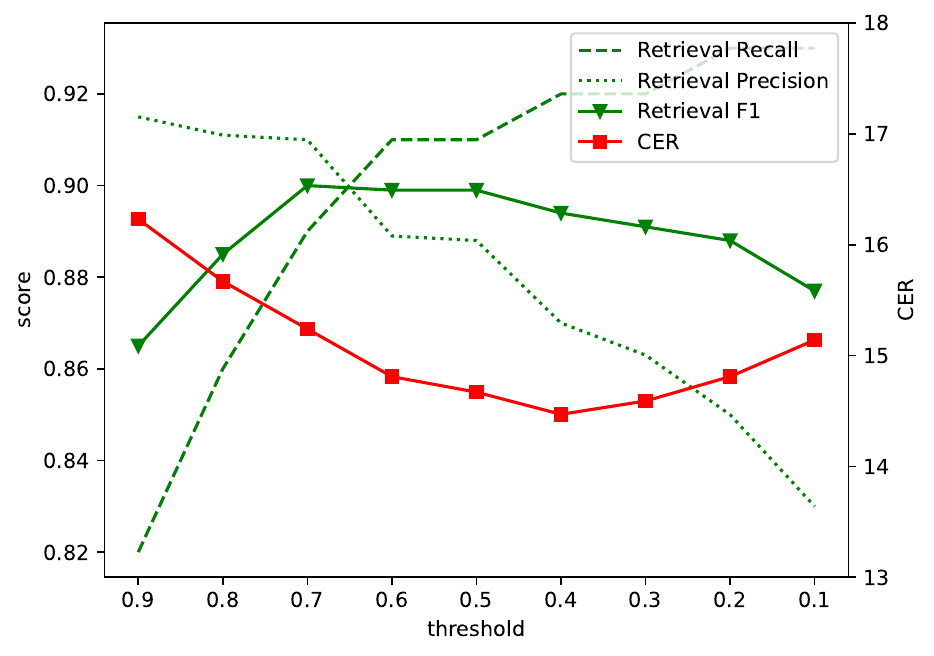}
\caption{Error Correction  
CER at different retrieval threshold.}
\label{figure:04}
\end{figure}

\subsection{Entity Rejection}
\label{entityreject}

Both steps of our method have the capability of entity rejection. In step 1, entity retrieval, we can filter out content with low similarity. In step 2, generative annotation, we can also reject entities by generating the symbol "<empty>". Since step 2 has the ability to reject correction, so we can allow more candidate entities retrieved in the step 1, without worrying about the accumulation of errors brought to step 2. 



Our retrieval step is noise-tolerant and does not require precisely accurate retrieval results. Figure \ref{figure:04} presents different F1 scores in the retrieval step based on different filter thresholds we set. According to the figure, the highest retrieval F1 score does not result in the best correction performance. Instead, higher recall and lower precision scores lead to the best correction accuracy, indicating that our correction method is fault-tolerant in terms of the retrieval results. 

\begin{figure}
\centering
\includegraphics[height=3.0cm,width=11.0cm]{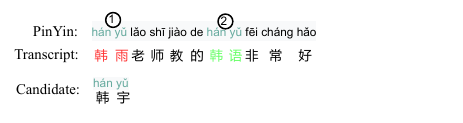}

\caption{This case contains two pieces of text that sound the same as the candidate entities, but only one of them needs correction. The first "\begin{CJK}{UTF8}{gbsn}{韩雨}\end{CJK}" is a person's name that should be corrected to "\begin{CJK}{UTF8}{gbsn}{韩宇}\end{CJK}". Although the pronunciation of the second piece of text "\begin{CJK}{UTF8}{gbsn}{韩语}\end{CJK}" is the same as the candidate entity, it does not require correction.}
\label{figure:EntityReject}
\end{figure}

If multiple words/phrases sound similar in the transcript but only one of them needs correction, phonetic-level similarity-based algorithms can hardly distinguish which one to correct. As shown in Figure \ref{figure:EntityReject}, the candidate entity "\begin{CJK}{UTF8}{gbsn}{韩宇}\end{CJK}" is a person's name, but in the transcript, there are two pieces of text that sound the same as the candidate entity, "\begin{CJK}{UTF8}{gbsn}{韩雨}\end{CJK}" (a person name but using a different Chinese character) and "\begin{CJK}{UTF8}{gbsn}{韩语}\end{CJK}" (means Korean language). We need to correct the first piece of text "\begin{CJK}{UTF8}{gbsn}{韩雨}\end{CJK}" (another person name) without correcting the second phonetically identical word "\begin{CJK}{UTF8}{gbsn}{韩语}\end{CJK}" (Korean language). PED-NEC corrects both pieces of text, leading to over-correction. Interestingly, our generative approach only corrects the first word and skips the second one, indicating that our model has the ability to determine which of the phonetically-similar words need correction. 

We believe such capability benefits from the use of the generative model's language model ability, which allows the model to learn that the candidate entity might be a person's name. Since the first piece of text is more like a person's name while the second piece of text is not relevant, so the model only corrects the first piece of text. According to the heatmap shown in Figure \ref{figure:heatmap}, the annotated result, which needs to be corrected, pays a lot of attention to the contex as well.


It should be noted that as shown in Table \ref{tab:ECform}, our method has the ability to annotate multiple incorrect forms of a candidate entity in one piece of ASR transcript.

\subsection{Corrupted Entity Detection}

When the number of entities increases, PED-NEC requires an important preliminary module, which is called Corrupted Entity Detection (CED). CED can detect NEs that are incorrectly transcribed in the ASR transcript, allowing PED-NEC to correct only these detected results. This effectively avoids over-correcting some words that are phonetically similar but are actually not entities. However, in our method, we did not use this preliminary module. We believe our GL method already possesses the capability of CED. Our training goal is to generate corrupted entities based on speech segment and prompt, indicating our model already has the capability of CED. This is also a potential advantage of our proposed generative correction method: it simultaneously performs CED and correction.

\section{Conclusion}

This article focuses on post-editing ASR errors and proposes a new generative error correction method to address a drawback of PED-NEC: fails to correct entities when the word form of the to-be-corrected text differs greatly from that of the ground-truth entity. Our method uses a generative approach to annotate to-be-corrected text in transcript based on speech segment, candidate entity and ASR transcript, and make replacement accordingly. This generative method is flexible and applicable to various entity correction scenarios. Our method also has the ability of entity rejection, an ability to decide when correction is not required. This ability allows more candidate entities in entity retrieval and further improves correction performance. Our method outperforms the baseline (PED-NE) on the open-source Aishell test set and our BuzzWord test set, no matter using the open-source Whisper or commercial ASR engines, thus demonstrating generalizability of our method.



 \section*{Limitations}

 Our method employs a Post-Correction strategy, so latency is a concern. Our method consists of two steps: NE retrieval and NE correction. Although our generative correction method only annotates to-be-corrected text, resulting in minimal time consumption, entity retrieval can become significantly time-consuming when there are many entities in the datastore. In such cases, on one hand, we can replace the retrieval with PED, which is the previously mentioned PED+GL method to reduce the overall latency; on the other hand, in the future, we plan to turn our retrieval approach into vector search, which can significantly accelerate speed through the use of existing mature vector search engines.



\begin{thebibliography}{29}
\expandafter\ifx\csname natexlab\endcsname\relax\def\natexlab#1{#1}\fi

\bibitem[{Bruguier(2019)}]{8682441}
Antoine~et.al Bruguier. 2019.
\newblock \href {https://doi.org/10.1109/ICASSP.2019.8682441} {Phoebe: Pronunciation-aware contextualization for end-to-end speech recognition}.
\newblock In \emph{ICASSP}, pages 6171--6175.

\bibitem[{Bu et~al.(2017)Bu, Du, Na, Wu, and Zheng}]{8384449}
Hui Bu, Jiayu Du, Xingyu Na, Bengu Wu, and Hao Zheng. 2017.
\newblock \href {https://doi.org/10.1109/ICSDA.2017.8384449} {Aishell-1: An open-source mandarin speech corpus and a speech recognition baseline}.
\newblock In \emph{2017 20th Conference of the Oriental Chapter of the International Coordinating Committee on Speech Databases and Speech I/O Systems and Assessment (O-COCOSDA)}, pages 1--5.

\bibitem[{Chen et~al.(2022)Chen, Xu, Wang, Xie, Zhang, and Huang}]{chen2022aishell}
Boli Chen, Guangwei Xu, Xiaobin Wang, Pengjun Xie, Meishan Zhang, and Fei Huang. 2022.
\newblock Aishell-ner: Named entity recognition from chinese speech.
\newblock In \emph{2022 IEEE International Conference on Acoustics, Speech and Signal Processing (ICASSP)}.

\bibitem[{Chorowski et~al.(2014)Chorowski, Bahdanau, Cho, and Bengio}]{6af3452a28a04980b2b8f5eb48730d36}
Jan Chorowski, Dzmitry Bahdanau, Kyunghyun Cho, and Yoshua Bengio. 2014.
\newblock End-to-end continuous speech recognition using attention-based recurrent nn: First results.
\newblock In \emph{NIPS}.

\bibitem[{Devlin et~al.(2019)Devlin, Chang, Lee, and Toutanova}]{devlin2019bert}
Jacob Devlin, Ming-Wei Chang, Kenton Lee, and Kristina Toutanova. 2019.
\newblock \href {http://arxiv.org/abs/1810.04805} {Bert: Pre-training of deep bidirectional transformers for language understanding}.
\newblock \emph{NAACL}.

\bibitem[{Dutta et~al.(2020)Dutta, Jain, and Maheshwari}]{Wang2020ASREC}
Samrat Dutta, Shreyansh Jain, and Ayush Maheshwari. 2020.
\newblock \href {https://api.semanticscholar.org/CorpusID:226202635} {Asr error correction with augmented transformer for entity retrieval}.
\newblock In \emph{Interspeech}.

\bibitem[{et.al(2020{\natexlab{a}})}]{Garg2020}
Abhinav~Garg et.al. 2020{\natexlab{a}}.
\newblock \href {https://doi.org/10.21437/Interspeech.2020-3174} {{Hierarchical Multi-Stage Word-to-Grapheme Named Entity Corrector for Automatic Speech Recognition}}.
\newblock In \emph{Proc. Interspeech}, pages 1793--1797.

\bibitem[{et.al(2021)}]{chen2021gigaspeech}
Guoguo~Chen et.al. 2021.
\newblock \href {http://arxiv.org/abs/2106.06909} {Gigaspeech: An evolving, multi-domain asr corpus with 10,000 hours of transcribed audio}.

\bibitem[{et.al(2020{\natexlab{b}})}]{jain2020contextual}
Mahaveer~Jain et.al. 2020{\natexlab{b}}.
\newblock \href {http://arxiv.org/abs/2006.03411} {Contextual rnn-t for open domain asr}.
\newblock \emph{Interspeech}.

\bibitem[{et.al(2023{\natexlab{a}})}]{leng2023softcorrect}
Yichong~Leng et.al. 2023{\natexlab{a}}.
\newblock \href {http://arxiv.org/abs/2212.01039} {Softcorrect: Error correction with soft detection for automatic speech recognition}.

\bibitem[{et.al(2020{\natexlab{c}})}]{liu-etal-2020-multilingual-denoising}
Yinhan~Liu et.al. 2020{\natexlab{c}}.
\newblock \href {https://doi.org/10.1162/tacl_a_00343} {Multilingual denoising pre-training for neural machine translation"}.
\newblock pages 726--742.

\bibitem[{et.al(2023{\natexlab{b}})}]{zhang2023google}
Yu~Zhang et.al. 2023{\natexlab{b}}.
\newblock \href {http://arxiv.org/abs/2303.01037} {Google usm: Scaling automatic speech recognition beyond 100 languages}.

\bibitem[{Graves(2012)}]{Graves_2012}
Alex Graves. 2012.
\newblock Sequence transduction with recurrent neural networks.
\newblock \emph{International Conference on Machine Learning,International Conference on Machine Learning}.

\bibitem[{Graves and Jaitly(2014)}]{Graves_Jaitly_2014}
Alex Graves and Navdeep Jaitly. 2014.
\newblock Towards end-to-end speech recognition with recurrent neural networks.
\newblock \emph{International Conference on Machine Learning,International Conference on Machine Learning}.

\bibitem[{Guo et~al.(2019)Guo, Sainath, and Weiss}]{guo2019spelling}
Jinxi Guo, Tara~N. Sainath, and Ron~J. Weiss. 2019.
\newblock \href {http://arxiv.org/abs/1902.07178} {A spelling correction model for end-to-end speech recognition}.
\newblock \emph{ICASSP}.

\bibitem[{Huber et~al.(2021)Huber, Hussain, Stüker, and Waibel}]{huber2021instant}
Christian Huber, Juan Hussain, Sebastian Stüker, and Alexander Waibel. 2021.
\newblock \href {http://arxiv.org/abs/2107.02268} {Instant one-shot word-learning for context-specific neural sequence-to-sequence speech recognition}.

\bibitem[{Le(2021)}]{le2021contextualized}
Duc~et.al Le. 2021.
\newblock Contextualized streaming end-to-end speech recognition with trie-based deep biasing and shallow fusion.
\newblock \emph{Interspeech}.

\bibitem[{Leng et~al.(2022{\natexlab{a}})Leng, Tan, Wang, Zhu, Xu, Liu, Liu, Qin, Li, Lin, and Liu}]{leng2022fastcorrect2}
Yichong Leng, Xu~Tan, Rui Wang, Linchen Zhu, Jin Xu, Wenjie Liu, Linquan Liu, Tao Qin, Xiang-Yang Li, Edward Lin, and Tie-Yan Liu. 2022{\natexlab{a}}.
\newblock \href {http://arxiv.org/abs/2109.14420} {Fastcorrect 2: Fast error correction on multiple candidates for automatic speech recognition}.

\bibitem[{Leng et~al.(2022{\natexlab{b}})Leng, Tan, Zhu, Xu, Luo, Liu, Qin, Li, Lin, and Liu}]{leng2022fastcorrect}
Yichong Leng, Xu~Tan, Linchen Zhu, Jin Xu, Renqian Luo, Linquan Liu, Tao Qin, Xiang-Yang Li, Ed~Lin, and Tie-Yan Liu. 2022{\natexlab{b}}.
\newblock \href {http://arxiv.org/abs/2105.03842} {Fastcorrect: Fast error correction with edit alignment for automatic speech recognition}.

\bibitem[{Ma et~al.(2023)Ma, Gales, Knill, and Qian}]{Ma2023NbestTR}
Rao Ma, Mark John~Francis Gales, Kate Knill, and Mengjie Qian. 2023.
\newblock \href {https://api.semanticscholar.org/CorpusID:257254986} {N-best t5: Robust asr error correction using multiple input hypotheses and constrained decoding space}.
\newblock \emph{Proc. Interspeech}, abs/2303.00456.

\bibitem[{Panayotov et~al.(2015)Panayotov, Chen, Povey, and Khudanpur}]{7178964}
Vassil Panayotov, Guoguo Chen, Daniel Povey, and Sanjeev Khudanpur. 2015.
\newblock \href {https://doi.org/10.1109/ICASSP.2015.7178964} {Librispeech: An asr corpus based on public domain audio books}.
\newblock In \emph{2015 IEEE International Conference on Acoustics, Speech and Signal Processing (ICASSP)}, pages 5206--5210.

\bibitem[{Pundak et~al.(2018)Pundak, Sainath, Prabhavalkar, Kannan, and Zhao}]{pundak2018deep}
Golan Pundak, Tara~N. Sainath, Rohit Prabhavalkar, Anjuli Kannan, and Ding Zhao. 2018.
\newblock \href {http://arxiv.org/abs/1808.02480} {Deep context: end-to-end contextual speech recognition}.
\newblock \emph{SLT}.

\bibitem[{Radford et~al.(2022)Radford, Kim, Xu, Brockman, McLeavey, and Sutskever}]{radford2022robust}
Alec Radford, Jong~Wook Kim, Tao Xu, Greg Brockman, Christine McLeavey, and Ilya Sutskever. 2022.
\newblock \href {http://arxiv.org/abs/2212.04356} {Robust speech recognition via large-scale weak supervision}.

\bibitem[{Raghuvanshi(2019)}]{raghuvanshi-etal-2019-entity}
Arushi~et.al Raghuvanshi. 2019.
\newblock Entity resolution for noisy {ASR} transcripts.

\bibitem[{Wang et~al.(2023)Wang, Liu, Li, and Zhao}]{wang2023improving}
Xiaoqiang Wang, Yanqing Liu, Jinyu Li, and Sheng Zhao. 2023.
\newblock \href {http://arxiv.org/abs/2302.11192} {Improving contextual spelling correction by external acoustics attention and semantic aware data augmentation}.
\newblock \emph{Proc. ICASSP}.

\bibitem[{Wang et~al.(2024)Wang, Wang, Yan, Lin, and Chen}]{wang2024dancer}
Yi-Cheng Wang, Hsin-Wei Wang, Bi-Cheng Yan, Chi-Han Lin, and Berlin Chen. 2024.
\newblock \href {http://arxiv.org/abs/2403.17645} {Dancer: Entity description augmented named entity corrector for automatic speech recognition}.

\bibitem[{Yadav et~al.(2020)Yadav, Ghosh, Yu, and Shah}]{yadav2020end}
Hemant Yadav, Sreyan Ghosh, Yi~Yu, and Rajiv~Ratn Shah. 2020.
\newblock End-to-end named entity recognition from english speech.
\newblock \emph{arXiv preprint arXiv:2005.11184}.

\bibitem[{Zhang and Huang(2020)}]{zhang-etal-2020-spelling}
Shaohua Zhang and Haoran~et.al Huang. 2020.
\newblock \href {https://doi.org/10.18653/v1/2020.acl-main.82} {Spelling error correction with soft-masked {BERT}}.
\newblock In \emph{ACL}, pages 882--890, Online.

\bibitem[{Zhang et~al.(2019)Zhang, Lei, and Yan}]{zhang2019automatic}
Shiliang Zhang, Ming Lei, and Zhijie Yan. 2019.
\newblock \href {http://arxiv.org/abs/1904.10045} {Automatic spelling correction with transformer for ctc-based end-to-end speech recognition}.

\end{thebibliography}
\bibliographystyle{acl_natbib}

\appendix

\section{Appendix}
\label{sec:appendix}
\subsection{BuzzWord Test Set}
To better demonstrate the generalizability of our method, we construct a new test set. We collect 50 buzzwords in Chinese from different areas (including tech, entertainment, social news, etc.) since January 2023. For each buzzwords, as shown in Table \ref{tab:testset}, we collect 5 videos (i.e. 5 speakers) on Bilibili\footnote{http://bilibili.com/} or YouTube\footnote{https://www.youtube.com/}.In every video, we extract two sentences that contains the buzzwords as positive examples and 4 sentences that does not contain the buzzword as negative examples. Finally, we get a 1500-sentence test set with 500 positive examples and 1000 negative examples. The duration of the audio recordings ranges from 5 to 15 seconds.

\begin{table}
\centering
\begin{tabular}{l|c|c|c}
\hline
 {} & Speaker & Positive & Negative\\
\hline
  & S1 & 2 &  4 \\
   \cline{2-4}
  & S2 & 2 &  4 \\
  \cline{2-4}
Entity  & S3 & 2 &  4 \\
\cline{2-4}
  & S4 & 2 &  4 \\
  \cline{2-4}
  & S5 & 2 &  4 \\
\hline
\end{tabular}

\caption{The details of creating one entity's positive and negative samples in our challenge test set.}
\label{tab:testset}
\end{table}

\subsection{Entity Info}
\begin{figure}
\centering
\setlength{\belowcaptionskip}{-0.4cm}
\includegraphics[height=5.5cm,width=7.5cm]{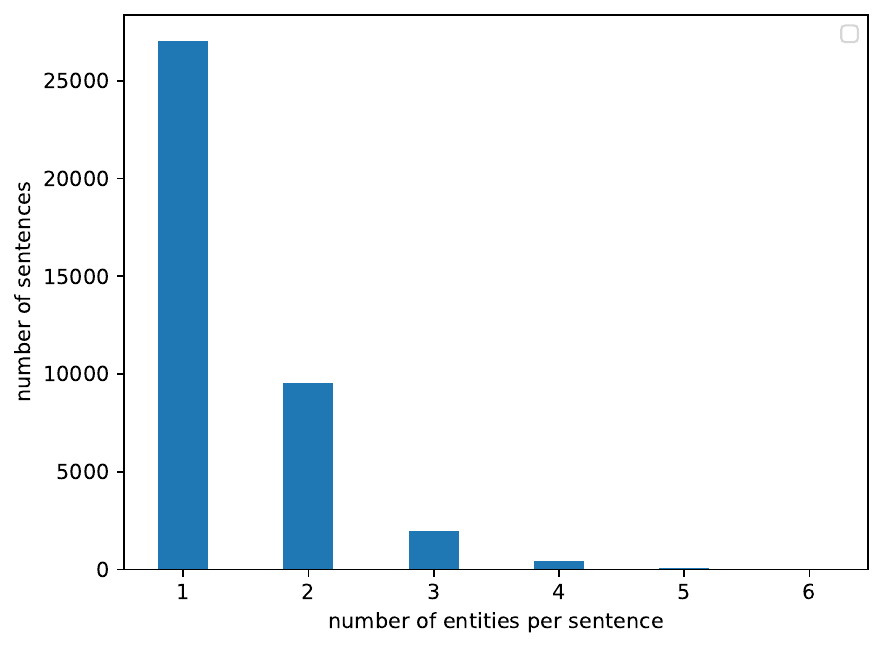}

\caption{Histogram of the distribution of entity counts in training data}
\label{figure:entityinfo}
\end{figure}
We also analyze the number of entity occurrences in the training data, as shown in Figure \ref{figure:entityinfo}. We found that the majority of training data only contains one entity per sentence, with a minority of sentences containing two entities. To address the correction of more entities, it is necessary to build a more diverse training dataset.
\subsection{Experimental Details}
\label{expdetails}

We are grateful for the work of \citet{wang2024dancer}. The baseline method PED-NEC was implemented entirely according to their open-source code\footnote{https://github.com/Amiannn/Dancer}. We used their bert-base CED method as the preliminary module for error correction in PED-NEC.
\begin{table*}
\centering
\begin{tabular}{l|cccc}
\hline
 & \multicolumn{4}{c}{\textbf{BuzzWord  Test Set (\%)}}\\
\cline{2-5}
 \textbf{Model} &  & NNE & NE  & NE\\
 & CER$\downarrow$ & CER$\downarrow$ &CER$\downarrow$ &  Recall$\uparrow$\\
\hline
iFlytek & 12.48 & \textbf{11.18} & 56.46 & 19.18   \\
\hline
PED-NEC & 12.29	& 11.41	& 45.26	& 46.42  \\
\hline
SS+GL &\textbf{11.28}	&11.19	&\textbf{14.09}	&\textbf{81.71}	 \\
\hline
\hline
Amazon & 25.88 & \textbf{24.40} & 73.67 & 9.84   \\
\hline
PED-NEC & 25.33	& 24.46	& 59.53	& 39.36  \\
\hline
SS+GL &\textbf{23.23}	&24.40	&\textbf{19.42}	&\textbf{80.02}	 \\
\hline
\end{tabular}
\caption{The commercial engine experiment results of our error correction method on the BuzzWord test set.}
\label{table:buzzwordcommercial}
\end{table*}
It should be noted that their CED module did not perform well in our BuzzWord test set, resulting in many corrupted entities not being detected. Consequently, we ultimately used the PED-NEC method without CED on the BuzzWord test set. We adjusted different similarity thresholds and selected the overall best CER result as the final outcome for PED-NEC.
\subsection{Correction for Commercial Engine}

To better verify the generalizability of our method, we also conducted error correction comparative experiments on the results of commercial engines (iFlytek\footnote{https://www.xfyun.cn/services/lfasr} and Amazon\footnote{https://aws.amazon.com/transcribe}). The results of the experiments on the BuzzWord test set showed that our method still significantly outperforms the PED-NEC method.

\subsection{Correction Cases}

\begin{table*}
\centering
\begin{tabular}{c|c}
\hline
No. & Result \\
\hline 
1 & \makecell*[l]{\begin{CJK}{UTF8}{gbsn}{Ref: 我看到咱们的电影《\textcolor{green}{茶啊二中}  (chá ā èr zhōng)》的时候...}\end{CJK}\\ 
\begin{CJK}{UTF8}{gbsn}{ASR: 我看到咱们的电影《\textcolor{red}{茶二中} (chá èr zhōng)》的时候...}\end{CJK} \\ 
\begin{CJK}{UTF8}{gbsn}{PED-NEC:我看到咱们的电影《\textcolor{red}{茶二中} (chá èr zhōng)》的时候...}\end{CJK} \\
\begin{CJK}{UTF8}{gbsn}{Ours: 我看到咱们的电影《\textcolor{green}{茶啊二中} (chá ā èr zhōng)》的时候...}\end{CJK} \\
\begin{CJK}{UTF8}{gbsn}{Explanation: "啊 (ā)" is a common filler word in Chinese. Perhaps the ASR system}\end{CJK} \\
\begin{CJK}{UTF8}{gbsn}{deliberately skips the word as a result of disfluency detection, or simply fails to transcribe}\end{CJK}\\
\begin{CJK}{UTF8}{gbsn}{the word.}\end{CJK}\\}\\
\hline
2& \makecell*[l]{\begin{CJK}{UTF8}{gbsn}{Ref: 但是我会认为它是真正促成《\textcolor{green}{苍兰诀} (cāng lán jué)》爆火的关键}\end{CJK}\\ 
\begin{CJK}{UTF8}{gbsn}{ASR: 但是我会认为他是真正促成\textcolor{red}{他在这} (tā zài zhè)爆火的关键}\end{CJK} \\ 
\begin{CJK}{UTF8}{gbsn}{PED-NEC: 但是我会认为他是真正促成\textcolor{red}{他在这} (tā zài zhè)爆火的关键}\end{CJK} \\
\begin{CJK}{UTF8}{gbsn}{Ours: 但是我会认为他是真正促成\textcolor{green}{苍兰诀} (cāng lán jué)爆火的关键}\end{CJK}  \\
\begin{CJK}{UTF8}{gbsn}{Explanation: "苍兰诀" is an OOV word to the ASR system. In addition, the background}\end{CJK} \\
\begin{CJK}{UTF8}{gbsn}{music in the audio makes it even harder to transcribe the entity. As a result, the transcribed} \end{CJK} \\
\begin{CJK}{UTF8}{gbsn}{result is total different from the ground-truth in terms of pronunciation.}\end{CJK}\\}\\
\hline
3 & \makecell*[l]{\begin{CJK}{UTF8}{gbsn}{Ref: \textcolor{green}{猴痘}患者可能性其实还是蛮低的，另外\textcolor{green}{猴痘} (hóu dòu)病毒它其实...}\end{CJK}\\ 
\begin{CJK}{UTF8}{gbsn}{ASR: \textcolor{red}{猴动}患者可能性其实还是蛮低的另外\textcolor{red}{猴动} (hóu dòng)病毒它其实...}\end{CJK} \\ 
\begin{CJK}{UTF8}{gbsn}{PED-NEC:\textcolor{red}{猴动}患者可能性其实还是蛮低的另外\textcolor{red}{猴动} (hóu dòng)病毒它其实...}\end{CJK} \\
\begin{CJK}{UTF8}{gbsn}{Ours: \textcolor{green}{猴痘}患者可能性其实还是蛮低的另外\textcolor{green}{猴痘} (hóu dòu)病毒它其实...}\end{CJK} \\
\begin{CJK}{UTF8}{gbsn}{Explanation: A mistranscription of "猴痘 (hóu dòu)} to phonetically-similar words "猴动\end{CJK} \\
\begin{CJK}{UTF8}{gbsn}{(hóu dòng)."}\end{CJK}\\}\\
\hline
4 & \makecell*[l]{\begin{CJK}{UTF8}{gbsn}{Ref: 主要就是focus在我们如果在本地利用我们\textcolor{green}{ChatGLM-6B}做一个本地的部署。}\end{CJK}\\ 
\begin{CJK}{UTF8}{gbsn}{ASR: 主要就是Focus在我们如果在本地利用我们\textcolor{red}{ChestJM6B}做一个本地的部署
}\end{CJK} \\ 
\begin{CJK}{UTF8}{gbsn}{PED-NEC: 主要就是Focus在我们如果在本地利用我们\textcolor{red}{ChestJM6B}做一个本地的部署}\end{CJK} \\
\begin{CJK}{UTF8}{gbsn}{Ours: 主要就是Focus在我们如果在本地利用我们\textcolor{green}{ChatGLM-6B}做一个本地的部署}\end{CJK}  \\
\begin{CJK}{UTF8}{gbsn}{Explanation: A mistranscription of "ChatGLM" to "ChestJM".}\end{CJK}\\}\\
\hline
5 & \makecell*[l]{\begin{CJK}{UTF8}{gbsn}{Ref:在I/O大会上，\textcolor{green}{ChatGPT}和新必应的竞争对手Bard经历了大幅更新。}\end{CJK}\\ 
\begin{CJK}{UTF8}{gbsn}{ASR: 在IO大会上 \textcolor{red}{Check GPT}和新必应的竞争对手Bard经历了大幅更新}\end{CJK} \\ 
\begin{CJK}{UTF8}{gbsn}{PED-NEC: 在IO大会上 \textcolor{red}{Check GPT}和新必应的竞争对手Bard经历了大幅更新
}\end{CJK} \\
\begin{CJK}{UTF8}{gbsn}{Ours: 在IO大会上 \textcolor{green}{ChatGPT}和新必应的竞争对手Bard经历了大幅更新}\end{CJK}\\
\begin{CJK}{UTF8}{gbsn}{Explanation: A mistranscription of "ChatGPT" to "Check GPT".}\end{CJK}\\}\\
\hline
6 & \makecell*[l]{\begin{CJK}{UTF8}{gbsn}{Ref: 所以这期视频呢带大家看的就是在这次发布的\textcolor{green}{Matebook D 16}。}\end{CJK}\\ 
\begin{CJK}{UTF8}{gbsn}{ASR: 所以这期视频呢带大家看的就是在这次发布的\textcolor{red}{matebook第16} (dì)}\end{CJK} \\
\begin{CJK}{UTF8}{gbsn}{PED-NEC: 所以这期视频呢带大家看的就是在这次发布的\textcolor{green}{Matebook D 16}\textcolor{red}{ook第16} (dì)}\end{CJK} \\
\begin{CJK}{UTF8}{gbsn}{Ours: 所以这期视频呢带大家看的就是在这次发布的\textcolor{green}{Matebook D 16}}\end{CJK}\\
\begin{CJK}{UTF8}{gbsn}{Explanation: A mistranscription of English letter "D" to Chinese word "第 (dì)", as they }\end{CJK} \\
\begin{CJK}{UTF8}{gbsn}{share similar pronunciations.}\end{CJK}\\}\\
\hline
\end{tabular}
\caption{More examples of comparing PED-NEC and our method when the word form of the transcribed entity results and the word form of the entity are different.}
\label{table:casestudycontinue}
\end{table*}

\end{document}